# *NPtool*, a detector of English noun phrases *

**Atro Voutilainen**
Research Unit for Computational Linguistics
P.O. Box 4 (Keskuskatu 8)
FIN-00014 University of Helsinki
Finland
E-mail: avoutila@ling.Helsinki.FI

## Abstract

*NPtool* is a fast and accurate system for extracting noun phrases from English texts for the purposes of e.g. information retrieval, translation unit discovery, and corpus studies. After a general introduction, the system architecture is presented in outline. Then follows an examination of a recently written Constraint Syntax. Section 6 reports on the performance of the system.

## 1 Introduction

This paper outlines *NPtool*, a noun phrase detector. At the heart of this modular system is **reductionistic word-oriented morphosyntactic analysis** that expresses head–modifier dependencies. Previous work on this approach, largely based on Karlsson's original proposal [Karlsson, 1990], is documented in [Karlsson *et al.*, forthcoming]. Let us summarise a few key features of this style of analysis.

• As most parsing frameworks, also the present style of analysis employs a lexicon and a grammar. What may distinguish the present approach from most other frameworks is the considerable degree of attention we pay to the **morphological and lexical description**: morphological analysis is based on an extensive and detailed description that employs inflectional and central derivational categories as well as other morphosyntactic features that can be useful for stating syntactic generalisations. In this way a carefully built and informative lexicon facilitates the construction of accurate, wide-coverage parsing grammars.

• We use **tags** to encode morphological distinctions, part of speech, and also syntactic information; for instance:

```
I       PRON     @HEAD
see     V PRES   @VERB
a       ART      @>N
bird    N        @HEAD
.       FULLSTOP
```

In this type of analysis, each word is provided with tags indicating e.g. part of speech, inflection, derivation, and syntactic function.

• Morphological and syntactic descriptions are based on **hand-coded linguistic rules** rather than on corpus-based statistical models. They employ structural categories that can be found in descriptive grammars, e.g. [Quirk, Greenbaum, Leech and Svartvik, 1985].

Regarding the at times heated methodological debate on whether statistical or rule-based information is to be preferred in grammatical analysis of running text (cf. e.g. [Sampson, 1987a; Taylor, Grover and Briscoe, 1989; Church, 1992]), we do not object to probabilistic methods in principle; nevertheless, it seems to us that rule-based descriptions are preferable because they can provide for more accurate and reliable analyses than current probabilistic systems, e.g. part-of-speech taggers [Voutilainen, Heikkilä and Anttila, 1992; Voutilainen, forthcoming a].[1] Proba-

---

*This paper, published in the *Proceedings of Workshop on Very Large Corpora* held June 22 1993 at Ohio State University, is based on a longer manuscript with the same title. The development of ENGCG was supported by TEKES, the Finnish Technological Development Center, and a part of the work on Finite-state syntax has been supported by the Academy of Finland.

[1]Consider for instance the question posed in [Church, 1992] whether lexical probabilities contribute more to morphological or part-of-speech disambiguation than

bilistic or heuristic techniques may still be a useful add-on to linguistic information, if potentially remaining ambiguities must be resolved – though with a higher risk of error.

• In the design of our grammar schemes, we have paid considerable attention to the question on the **resolvability of grammatical distinctions**. In the design of accurate parsers of running text, this question is very important: if the description abounds with distinctions that can be dependably resolved only with extrasyntactic knowledge[2], then either the ambiguities due to these distinctions remain to burden the structure-based parser (as well as the potential application based on the analysis), or a guess, i.e. a misprediction, has to be hazarded.

This descriptive policy brings with it a certain degree of **shallowness**; in terms of information content, a tag-based syntactic analysis is somewhere between morphological (e.g. part-of-speech) analysis and a conventional syntactic analysis, e.g. a phrase structure tree or a feature-based analysis. What we hope to achieve with this compromise in information content is the higher reliability of the proposed analyses. A superior accuracy could be considered as an argument for postulating a new, 'intermediary' level of computational syntactic description. For more details, see e.g. [Voutilainen and Tapanainen, 1993].

• Our grammar schemes are also *learnable*: according to double-blind experiments on manually assigning morphological descriptions, a 100% interjudge agreement is typical [Voutilainen, forthcoming].[3]

• The ability to parse running text is of a high priority. Not only a structurally motivated description is important; in the construction of the parsing grammars and lexica, attention should also be paid to corpus evidence. Often a grammar rule, as we express it in our parsing grammars, is formed as a generalisation 'inspired' by corpus observations; in this sense the parsing grammar is corpus-based. However, the description need not be *restricted* to the corpus observation: the linguist is likely to generalise over past experience, and this is not necessarily harmful – as long as the generalisations can also be validated against representative test corpora.

• At least in a practical application, a parsing grammar should assign **the best available analysis** to its input rather than leave many of the input utterances unrecognised e.g. as ill-formed. This does not mean that the concept of well-formedness is irrelevant for the present approach. Our point is simply: although we may consider some text utterance as deviant in one respect or another, we may still be interested in extracting as much information as possible from it, rather than ignore it altogether. To achieve this effect, the grammar rules should be used in such a manner[4] that no input becomes entirely rejected, although the rules as such may express categorical restrictions on what is possible or well-formed in the language.

• In our approach, parsing consists of two main kinds of operation:

1. Context-insensitive lookup of (alternative) descriptions for input words

2. Elimination of unacceptable or contextually illegitimate alternatives.

**Morphological analysis** typically corresponds to the lookup module: it produces the desired morphosyntactic analysis of the sentence, along with a number of inappropriate ones, by providing each word in the sentence with all conventional analyses as a list of alternatives. The grammar itself exerts the restrictions on permitted sequences of words and descriptors. In other words, syntactic analysis proceeds by way of **ambiguity resolution or disambiguation**: the parser eliminates ill-formed readings, and what 'survives' the grammar is the (syntactic) analysis of the input utterance. Since the input contains the desired analysis, no new structure will be built during syntactic analysis itself.

• Our grammars consist of **constraints** – partial distributional definitions of morphosyntactic categories, such as parts of speech or syntactic functions. Each constraint expresses a piecemeal linear-precedence generalisation about the language, and they are independent of each other. That is, the constraints can be applied in any order: a true grammar will produce the same analysis, whatever the order.

The grammarian is relatively free to select the level of abstraction at which (s)he is willing to express the distributional generalisation. In particular, also reference to very low-level categories is possible, and this makes for the accuracy of the parser: while the grammar will contain more or less abstract, feature-oriented rules, often it is also expedient to state further, more particular restrictions on more particular distributional classes, even at

---

context does. The ENGCG morphological disambiguator, which is entirely based on context rules, uniquely and correctly identifies more than 97% of all appropriate descriptions, and this is considerably more than the near-90% success rate achieved with lexical probabilities alone [Church, 1992]. Moreover, note that in all, the ENGCG disambiguator identifies more than 99.5% of all appropriate descriptions; only, some 2–3% of all analyses remain ambiguous and thus do not become uniquely identified. For more details, see [Voutilainen, forthcoming 1993].

[2]Witness, for instance, ambiguities due to adverbial attachment or modifier scope.

[3]The 95% interjudge agreement rate reported in [Church, 1992] probably indicates that in the case of debatable constructions, explicit descriptive conventions have not been consistently established. Only a carefully defined grammar scheme makes the evaluation of the accuracy of the parsing system a meaningful enterprise (see also [Sampson, 1987b]).

[4]e.g. by ranking the grammar rules in terms of compromisability

the word-form level. These 'smaller' rules do not contradict the more general rules; often it is simply the case that further restrictions can be imposed on smaller lexical classes[5] This flexibility in the grammar formalism greatly contributes to the accuracy of the parser [Voutilainen, forthcoming a; Voutilainen, forthcoming 1993].

## 2 Uses of a noun phrase parser

The recognition and analysis of subclausal structural units, e.g. noun phrases, is useful for several purposes. Firstly, a noun phrase detector can be useful for research purposes: automatic large-scale analysis of running text provides the linguist with better means to conduct e.g. quantitative studies over large amounts of text.

An accurate though somewhat superficial analysis can also serve as a 'preprocessor' prior to more ambitious, e.g. feature-based syntactic analysis. This kind of division of labour is likely to be useful for technical reasons. One major problem with e.g. unification-based parsers is parsing time. Now if a substantial part of the overall problem is resolved with more simple and efficient techniques, the task of the unification-based parser will become more manageable. In other words, the more expressive but computationally heavier machinery of e.g. the unification-based parser can be reserved entirely for the analysis of the descriptively hardest problems. The less complex parts of the overall problem can be tackled with more simple and efficient techniques.

Regarding production uses, even lower levels of analysis can be dearly useful. For instance, the detection of noun phrases can provide e.g. information management and retrieval systems with a suitable input for index term generation.

Noun phrases can also serve as translation units; for instance, [van der Eijk, 1993] suggests that noun phrases are more appropriate translation units than words or part-of-speech classes.

## 3 Previous work

This section consists of two subsections. Firstly, a performance-oriented survey of some related systems is presented. Then follows a more detailed presentation of ENGCG, a predecessor of the *NPtool* parser in an information retrieval system.

### 3.1 Related systems

So far, I have found relatively little documentation on systems whose success in recognising or parsing noun phrases has been reported. I am aware of three systems with some relevant evaluations.

Church's *Parts of speech* [Church, 1988] performs not only part-of-speech analysis, but it also identifies the most simple kinds of noun phrases – mostly sequences of determiners, premodifiers and nominal heads – by inserting brackets around them, e.g.

```
[A/AT former/AP top/NN aide/NN] to/IN
[Attorney/NP/NP General/NP/NP Edwin/NP/NP
Meese/NP/NP] interceded/VBD ...
```

The appendix in [Church, 1988] lists the analysis of a small text. The performance of the system on the text is quite interesting: of 243 noun phrase brackets, only five are omitted. – The performance of *Parts of speech* was also very good in part-of-speech analysis on the text: 99.5% of all words got the appropriate tag. The mechanism for noun phrase identification relies on the part-of-speech analysis; the part-of-speech tagger was more successful on the text than on an average; therefore the average performance of the system in noun phrase identification may not be quite as good as the figures in the appendix of the paper suggest.

Bourigault's LECTER [Bourigault, 1992] is a surface-syntactic analyser that extracts 'maximal-length noun phrases' – mainly sequences of determiners, premodifiers, nominal heads, and certain kinds of postmodifying prepositional phrases and adjectives – from French texts for terminology applications. The system is reported to recognise 95% of all maximal-length noun phrases (43,500 out of 46,000 noun phrases in the test corpus), but no figures are given on how much 'garbage' the system suggests as noun phrases. It is indicated, however, that manual validation is necessary.

Rausch, Norrback and Svensson [1992] have designed a noun phrase extractor that takes as its input part-of-speech analysed Swedish text, and inserts brackets around noun phrases. In the recognition of 'Nuclear Noun Phrases' – sequences of determiners, premodifiers and nominal heads – the system was able to identify 85.9% of all nuclear noun phrases in a text collection, some 6,000 words long in all, whereas some 15.7% of all the suggested noun phrases were false hits, i.e. the precision[6] of the system was 84.3%. The performance of a real application would probably be lower because potential misanalyses due to previous stages of analysis (morphological analysis and part-of-speech disambiguation, for instance) are not accounted for by these figures.

### 3.2 ENGCG and the SIMPR project

SIMPR, *Structured Information Management: Processing and Retrieval*, was a 64 person year ESPRIT II project (Nr. 2083, 1989–1992), whose objective was to develop new methods for the management and retrieval of large amounts of electronic texts. A central function of such a system is to recognise those words in the stored texts that represent it in a concise fashion – in short, **index terms**.

---

[5]Consider for instance the attachment of prepositional phrases in general and of *of*-phrases in particular.

[6]For definitions of the terms *recall* and *precision*, see Section 6.

Term indices created with traditional methods[7] are based on isolated, perhaps truncated words. These largely string-based statistical methods are somewhat unsatisfactory because many content identifiers consist of word sequences – compounds, head–modifier constructions, even simple verb – noun phrase sequences. One of the SIMPR objectives was also to employ more complex constructions, the recognition of which would require a shallow grammatical analysis. The Research Unit for Computational Linguistics at the University of Helsinki participated in this project, and ENGTWOL, a Twol-styled morphological analyser as well as ENGCG, a Constraint Grammar of English, were written 1989–1992 by Voutilainen, Heikkilä and Anttila [forthcoming]. The resultant SIMPR system is an improvement over previous systems [Smart (Ed.), forthcoming] – it is not only reasonably accurate, but also it operates on more complex constructions, e.g. postmodifying constructions and simple verb–object constructions.

There were also some persistent problems. The original plan was to use the output of the whole ENGCG parser for the indexing module. However, the last module of the three sequential modules in the ENGCG grammar, namely Constraint Syntax proper, was not used in the more mature versions of the indexing module – only lexical analysis and morphological disambiguation were applied. The omission of the syntactic analysis was mainly due to the somewhat high error rate (3–4% of all words lost the proper syntactic tag) and the high rate of remaining ambiguities (15–25% of all words remained syntactically ambiguous.

Here, we will not go into a detailed analysis of the problems[8], suffice it to say that the syntactic grammar scheme was unnecessarily ambitious for the relatively simple needs of the indexing application. One of the improvements in *NPtool* is a more optimal syntactic grammar scheme, as will be seen in Section 5.1.

## 4    *NPtool* in outline

In this section, the architecture of *NPtool* is presented in outline. Here is a flow chart of the system:

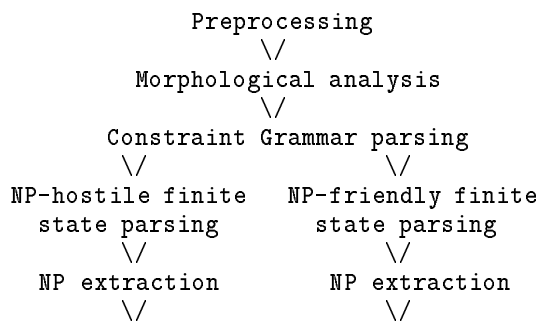

---

[7]See e.g. [Salton and McGill, 1983].

[8]See e.g. [Voutilainen, Heikkilä and Anttila, 1992] for details.

## Intersection of noun phrase sets

In the rest of this section, we will observe the analysis of the following sample sentence, taken from a car maintenance manual:

```
The inlet and exhaust manifolds are mounted
on opposite sides of the cylinder head, the
exhaust manifold channelling the gases to a
single exhaust pipe and silencer system.
```

### 4.1   Preprocessing and morphological analysis

The input ASCII text, preferably SGML-annotated, is subjected to a preprocessor that e.g. determines sentence boundaries, recognises fixed syntagms[9], normalises certain typographical conventions, and verticalises the text.

This preprocessed text is then submitted to morphological analysis. ENGTWOL, a morphological analyser of English, is a Koskenniemi-style morphological description that recognises all inflections and central derivative forms of English. The present lexicon contains some 56,000 word stems, and altogether the analyser recognises several hundreds of thousands of different word-forms. The analyser also employs a detailed parsing-oriented morphosyntactic description; the feature system is largely derived from [Quirk, Greenbaum, Leech and Svartvik, 1985]. Here is a small sample:

```
("<*the>"
  ("the" DET CENTRAL ART SG/PL (@>N)))
("<inlet>"
  ("inlet" N NOM SG))
("<and>"
  ("and" CC (@CC)))
("<exhaust>"
  ("exhaust" <SVO> V SUBJUNCTIVE VFIN (@V))
  ("exhaust" <SVO> V IMP VFIN (@V))
  ("exhaust" <SVO> V INF)
  ("exhaust" <SVO> V PRES -SG3 VFIN (@V))
  ("exhaust" N NOM SG))
("<manifolds>"
  ("manifold" N NOM PL))
```

All running-text word-forms are given on the left-hand margin, while all analyses are on indented lines of their own. The multiplicity of these lines for a word-form indicates morphological ambiguity.

For words not represented in the ENGTWOL lexicon, there is a 99.5% reliable utility that assigns ENGTWOL-style descriptions. These predictions are based on the form of the word, but also some heuristics are involved.

### 4.2   Constraint Grammar parsing

The next main stage in *NPtool* analysis is Constraint Grammar parsing. Parsing consists of two

---

[9]e.g. multiword prepositions and compounds

main phases: morphological disambiguation and Constraint syntax.

• **Morphological disambiguation.** The task of the morphological disambiguator is to discard all contextually illegitimate morphological readings in ambiguous cohorts. For instance, consider the following fragment:

```
("<a>"
  ("a" <Indef> DET CENTRAL ART SG (@>N)))
("<single>"
  ("single" <SVO> V IMP VFIN (@V))
  ("single" <SVO> V INF)
  ("single" A ABS))
```

Here an unambiguous determiner is directly followed by a three-ways ambiguous word, two of the analyses being verb readings, and one, an adjective reading. – A determiner is never followed by a verb[10]; one of the 1,100-odd constraints in the disambiguation grammar [Voutilainen, forthcoming a] expresses this fact about English grammar; so the verb readings of *single* are discarded here.

The morphological disambiguator seldom discards an appropriate morphological reading: after morphological disambiguation, 99.7–100% of all words retain the appropriate analysis. On the other hand, some 3–6% of all words remain ambiguous, e.g. *head* in this sentence. There is also an additional set of some 200 constraints – after the application of both constraint sets, 97–98% of all words become fully disambiguated, with an overall error rate of up to 0.4% [Voutilainen, forthcoming b]. The present disambiguator compares quite favourably with other known, typically probabilistic, disambiguators, whose maximum error rate is as high as 5%, i.e. some 17 times as high as that of the ENGCG disambiguator.

• **Constraint syntax.** After morphological disambiguation, the syntactic constraints are applied. In the *NPtool* syntactic description, all syntactic ambiguities are introduced directly in the lexicon, so no extra lookup module is needed. Like disambiguation constraints, syntactic constraints seek to discard all contextually illegitimate syntactic function tags. Here is the syntactic analysis of our sample sentence, as produced by the current parser. To save space, most of the morphological codes are omitted.

```
("<*the>"
  ("the" DET (@>N)))
("<inlet>"
  ("inlet" N (@>N @NH)))
("<and>"
  ("and" CC (@CC)))
("<exhaust>"
  ("exhaust" N (@>N)))
("<manifolds>"
```

---

---

```
  ("manifold" N (@NH)))
("<are>"
  ("be" V (@V)))
("<mounted>"
  ("mount" PCP2 (@V)))
("<on>"
  ("on" PREP (@AH)))
("<opposite>"
  ("opposite" A (@>N)))
("<sides>"
  ("side" N (@NH)))
("<of>"
  ("of" PREP (@N<)))
("<the>"
  ("the" DET (@>N)))
("<cylinder>"
  ("cylinder" N (@>N @NH)))
("<head>"
  ("head" V (@V))
  ("head" N (@NH)))
("<$,>")
("<the>"
  ("the" DET (@>N)))
("<exhaust>"
  ("exhaust" N (@>N)))
("<manifold>"
  ("manifold" N (@NH)))
("<channelling>"
  ("channel" PCP1 (@V)))
("<the>"
  ("the" DET (@>N)))
("<gases>"
  ("gas" N (@NH)))
("<to>"
  ("to" PREP (@AH)))
("<a>"
  ("a" DET (@>N)))
("<single>"
  ("single" A (@>N)))
("<exhaust>"
  ("exhaust" N (@>N)))
("<pipe>"
  ("pipe" N (@NH)))
("<and>"
  ("and" CC (@CC)))
("<silencer>"
  ("silencer" N (@>N)))
("<system>"
  ("system" N (@NH)))
("<$.>")
```

All syntactic-function tags are flanked with '@'. For instance, the tag '@>N' indicates that the word is a determiner or a premodifier of a nominal in the right-hand context (e.g. *the*). The second word, *inlet*, remains syntactically ambiguous due to a premodifier reading and a nominal head *@NH* reading – note that the ambiguity is structurally genuine, a coordination ambiguity. The tag *@V* is reserved for verbs and auxiliaries, cf. *are* as well as *mounted*. The syntactic description will be outlined below.

Pasi Tapanainen[11] has recently made a new implementation of the Constraint Grammar parser that performs morphological disambiguation and syntactic analysis at a speed of more than 1,000 words per second on a Sun SparcStation 10, Model 30.

## 4.3 Treatment of remaining ambiguities

The Constraint Grammar parser recognises only word-level ambiguities, therefore some of the traversals through an ambiguous sentence representation may be blatantly ill-formed.

*NPtool* eliminates locally unacceptable analyses by using a finite-state parser [Tapanainen, 1991][12] as a kind of 'post-processing module' that distinguishes between competing sentence readings. The parser employs a small finite-state grammar that I have written. The speed of the finite-state parser is comparable to that of the Constraint Grammar parser.

The finite-state parser produces all sentence readings that are in agreement with the grammar. Consider the following two adapted readings from the beginning of our sample sentence:

```
the/@>N inlet/@>N and/@CC exhaust/@>N
manifolds/@NH are/@V mounted/@V
on/@AH opposite/@>N sides/@NH
of/@N< the/@>N cylinder/@NH head/@V

the/@>N inlet/@>N and/@CC exhaust/@>N
manifolds/@NH are/@V mounted/@V
on/@AH opposite/@>N sides/@NH
of/@N< the/@>N cylinder/@>N head/@NH
```

The only difference is in the analysis of *cylinder head*: the first analysis reports *cylinder* as a noun phrase head which is followed by the verb *head*, while the second analysis considers *cylinder head* as a noun phrase. Now the last remaining problem is, how to deal with ambiguous analyses like these: should *cylinder* be reported as a noun phrase, or is *cylinder head* the unit to be extracted?

The present system provides all proposed noun phrase candidates in the output, but each with an indication of whether the candidate noun phrase is unambiguously analysed as such, or not. In this solution, I do not use all of the multiple analyses proposed by the finite-state parser. For each sentence, no more than two competing analyses are selected for further processing: one with the highest number of words as part of a maximally long noun phrase analysis, and the other with the lowest number of words as part of a maximally short noun phrase analysis.

This 'weighing' can be done during finite-state parsing: the formalism employs a mechanism for imposing penalties on regular expressions, e.g. on tags. A penalised reading is not discarded as ungrammatical, only the parser returns all accepted analyses in an order where the least penalised analyses are produced first and the 'worst' ones last.

Thus there is an 'NP-hostile' finite-state parser that penalises noun phrase readings; this would prefer the sentence reading with *cylinder/@NH head/@V*. The 'NP-friendly' parser, on the other hand, penalises all readings which are **not** part of a noun phrase reading, so it would prefer the analysis with *cylinder/@>N head/@NH*. Of all analyses, the selected two parses are maximally dissimilar with regard to NP-hood. The motivation for selecting maximally conflicting analyses in this respect is that a candidate noun phrase that is agreed upon as a noun phrase by the two finite-state parsers systems *just as it is* – neither longer nor shorter – is likely to be an unambiguously identified noun phrase. The comparison of the outputs of the two competing finite-state parsers is carried out during the extraction phase.

## 4.4 Extraction of noun phrases

An unambiguous sentence reading is a linear sequence of symbols, and extracting noun phrases from this kind of data is a simple pattern matching task.

In the present version of the system, I have used the *gawk* program that allows the use of regular expressions. With *gawk*'s *gsub* function, the boundaries of the longest non-overlapping expressions that satisfy the search key can be marked. If we formulate our search query as something like the following schematic regular expression:

```
[[M>N+ [CC M>N+]*]* HEAD
[N< [D/M>N+ [CC D/M>N+]*]* HEAD]*]
```

where

| | |
|---|---|
| '[' and ']' | are for grouping, |
| '+' | stands for one or more occurrences of its argument, |
| '*' | stands for zero or more occurrences of its argument, |
| 'M>N' | stands for premodifiers, |
| 'D/M>N' | stands for determiners and premodifiers, |
| 'HEAD' | stands for nominal heads except pronouns, |
| 'N<' | stands for prepositions starting a postmodifying prepositional phrase, |

and do some additional formatting and 'cleaning', the above two finite-state analyses will look like the following[13]:

```
the
```

[13] Note that the noun phrase heads are here given in the base form, hence the absence of the plural form of e.g. 'manifold'.

```
 np: inlet and exhaust manifold
are mounted on
 np: opposite side of the cylinder
head, the
 np: exhaust manifold
channelling the
 np: gas
to a
 np: single exhaust pipe
and
 np: silencer system
 .

the
 np: inlet and exhaust manifold
are mounted on
 np: opposite side of the cylinder head
, the
 np: exhaust manifold
channelling the
 np: gas
to a
 np: single exhaust pipe
and
 np: silencer system
 .
```

The proposed noun phrases are given on indented lines, each marked with the symbol 'np:'. The candidate noun phrases are then subjected to further routines: all candidate noun phrases with at least one occurrence in the output of both the NP-hostile and NP-friendly parsers are labelled with the symbol 'ok:', and the remaining candidates are labelled as uncertain, with the symbol '?:'. From the outputs given above, the following list can be produced:

```
ok: inlet and exhaust manifold
ok: exhaust manifold
ok: gas
ok: single exhaust pipe
ok: silencer system
?: opposite side of the cylinder
?: opposite side of the cylinder head
```

The linguistic analysis is relatively neutral as to what is to be extracted from it. Here we have concentrated on noun phrase extraction, but from this kind of input, also many other types of construction could be extracted, e.g. simple verb–argument structures.

## 5 The syntactic description

This section outlines the syntactic description that I have written for *NPtool* purposes. The ENGTWOL lexicon and the disambiguation constraints will not be described further in this paper; they have been documented extensively elsewhere (see the relevant articles in Karlsson & al. [forthcoming]).

According to the SIMPR experiences, the vast majority of index terms represent relatively few constructions. By far the most common construction is a nominal head with optional, potentially coordinated premodifiers and postmodifying prepositional phrases, typically *of* phrases. The remainder, less than 10%, consists almost entirely of relatively simple verb–NP patterns.

The syntactic description used in SIMPR employed some 30 dependency-oriented syntactic function tags, which differentiate (to some extent) between various kinds of verbal constructions, syntactic functions of nominal heads, and so on. Some of the ambiguity that survives ENGCG parsing is in part due to these distinctions [Anttila, forthcoming].

The relatively simple needs of an index term extraction utility on the one hand, and the relative abundance of distinctions in the ENGCG syntactic description on the other, suggest that a less distinctive syntactic description might be more optimal for the present purposes: a more shallow description would entail less remaining ambiguity without unduly compromising its usefulness e.g. for an indexing application.

### 5.1 Syntactic tags

I have designed a new syntactic grammar scheme that employs seven function tags. These tags capitalise on the opposition between noun phrases and other constructions on the one hand, and between heads and modifiers, on the other. Here we will not go into details; a gloss with a simple illustration will suffice.

• **@V** represents auxiliary and main verbs as well as the infinitive marker *to* in both finite and nonfinite constructions. For instance:

`She should/@V know/@V what to/@V do/@V.`

• **@NH** represents nominal heads, especially nouns, pronouns, numerals, abbreviations and *-ing*-forms. Note that of adjectival categories, only those with the morphological feature $<Nominal>$, e.g. *English*, are granted the *@NH* status: all other adjectives (and *-ed*-forms) are regarded as too unconventional nominal heads to be granted this status in the present description. An example:

`The English/@NH may like the conventional.`

• **@>N** represents determiners and premodifiers of nominals (the angle-bracket '>' indicates the direction in which the head is to be found). The head is the following nominal with the tag *@NH*, or a premodifier in between. An example:

`the/@>N fat/@>N butcher's/@>N wife`

• **@N<** represents prepositional phrases that unambiguously postmodify a preceding nominal head. Such unambiguously postmodifying constructions are typically of two types: (i) in the absence of certain verbs like 'accuse', postnominal *of*-phrases and (ii) preverbal NP—PP sequences, e.g.

```
The man in/@N< the moon had
a glass of/@N< ale.
```

Currently the description does not account for other types of postmodifier, e.g. postmodifying adjectives, numerals, other nominals, or clausal constructions.

● **@CC** and *@CS* represent co-ordinating and subordinating conjunctions, respectively:

```
Either/@CC you or/@CC I will go
if/@CS necessary.
```

● **@AH** represents the 'residual': adjectival heads, adverbials of various kinds, adverbs (also intensifiers), and also those of the prepositional phrases that cannot be dependably analysed as a postmodifier. An example is in order:

```
There/@AH have always/@AH been very/@AH
many people in/@AH this area.
```

### 5.2 Syntactic constraints

The syntactic grammar contains some 120 syntactic constraints. Like the morphological disambiguation constraints, these constraints are essentially negative partial linear-precedence definitions of the syntactic categories.

The present grammar is a partial expression of four general grammar statements:

1. Part of speech determines the order of determiners and modifiers.

2. Only likes coordinate.

3. A determiner or a modifier has a head.

4. An auxiliary is followed by a main verb.

We will give only one illustration of how these general statements can be expressed in Constraint Grammar. Let us give a partial paraphrase of the statement *Part of speech determines the order of determiners and modifiers*: 'A premodifying noun occurs closest to its head'. In other words, premodifiers from other parts of speech do not immediately follow a premodifying noun. Therefore, a noun in the nominative immediately followed by an adjective is not a premodifier. Thus a constraint in the grammar would discard the *@>N* tag of *Harry* in the following sample sentence, where *Harry* is directly followed by an unambiguous adjective:

```
("<*is>"
 ("be" <SVC/N> <SVC/A> V PRES SG3 (@V)))
("<*harry>"
 ("harry" <Proper> N NOM SG (@NH @>N)))
("<foolish>"
 ("foolish" A ABS (@AH)))
("<$?>")
```

We require that the noun in question is a nominative because premodifying nouns in the genitive can occur also before adjectival premodifiers; witness *Harry's* in *Harry's foolish self*.

### 5.3 Evaluation

The present syntax has been applied to large amounts of journalistic and technical text (newspapers, abstracts on electrical engineering, manuals on car maintenance, etc.), and the analysis of some 20,000–30,000 words has been proofread to get an estimate of the accuracy of the parser.

After the application of the *NPtool* syntax, some 93–96% of all words become syntactically unambiguous, with error rate of less than 1%[14].

To find out how much ambiguity remains at the sentence level, I also applied a 'NP-neutral' version[15] of the finite-state parser on a 25,500 word text from *The Grolier Electronic Encyclopaedia*. The results are given in Figure 1.

Figure 1: Ambiguity rates after finite-state parsing in a text of 1,495 sentences (25,500 words). **R** indicates the number of analyses per sentence, and **F** indicates the frequency of these sentences.

| R | F | R | F | R | F | R | F |
|---|-----|----|----|----|---|----|---|
| 1 | 960 | 6  | 19 | 12 | 6 | 32 | 2 |
| 2 | 304 | 7  | 3  | 14 | 3 | 48 | 2 |
| 3 | 54  | 8  | 28 | 16 | 5 | 64 | 1 |
| 4 | 93  | 9  | 3  | 24 | 1 | 72 | 1 |
| 5 | 4   | 10 | 3  | 28 | 1 |    |   |

Some 64% (960) of the 1,495 sentences became syntactically unambiguous, while only some 2% of all sentences analyses contain more than ten readings, the worst ambiguity being due to 72 analyses. This compares favourably with the ENGCG performance: after ENGCG parsing, 23.5% of all sentences remained ambiguous due to a number of sentence readings greater than the *worst* case in *NPtool* syntax.

## 6 Performance of *NPtool*

Various kinds of metrics can be proposed for the evaluation of a noun phrase extractor; our main metrics are **recall** and **precision**, defined as follows[16]:

● Recall: the ratio 'retrieved intended NPs'[17] / 'all intended NPs'

● Precision: the ratio 'all retrieved NPs' / 'retrieved intended NPs'

---

[14] This figure also covers errors due to previous stages of analysis.

[15] i.e. a parser which does not contain the mechanism for penalising or favouring noun phrase analyses; see Section 4.3 above.

[16] This definition also agrees with that used in Rausch & al. [1992].

[17] An 'intended NP' is the longest non-overlapping match of the search query given in extraction phase.

To paraphrase, a recall of less than 100% indicates that the system missed some of the desired noun phrases, while a precision of less than 100% indicates that the system retrieved something that is not regarded as a correct result.

The performance of the whole system has been evaluated against several texts from different domains. In all, the analysis of some 20,000 words has been manually checked.

If we wish to extract relatively complex noun phrases with optional coordination, premodifiers and postmodifiers (see the search query above in Section 4.4), we reach a recall of 98.5–100%, with a precision of some 95–98%.

As indicated in Section 4.4, the extraction utility annotates each proposed noun phrase as a 'sure hit' ('ok:') or as an 'uncertain hit' ('?:'). This distinction is quite useful for manual validation: approximately 95% of all superfluous noun phrase candidates are marked with the question mark.

## 7 Conclusion

In terms of accuracy, *NPtool* is probably one of the best in the field. In terms of speed, much remains to be optimised. Certainly the computationally most demanding tasks – disambiguation and parsing – are already carried out quite efficiently, but the more trivial parts of the system could be improved.

## 8 Acknowledgements

I wish to thank Krister Linden, Pasi Tapanainen and two anonymous referees for useful comments on an earlier version of this paper. The usual disclaimers hold.

APPENDIX

Here is given the *NPtool* analysis of a small sample from the CACM text collection. – Here is the input text:

```
The binary number system offers many
advantages over a decimal representation
for a high-performance, general-purpose
computer. The greater simplicity of a
binary arithmetic unit and the greater
compactness of binary numbers both
contribute directly to arithmetic
speed. Less obvious and perhaps more
important is the way binary addressing
and instruction formats can increase the
overall performance. Binary addresses
are also essential to certain powerful
operations which are not practical with
decimal instruction formats. On the other
hand, decimal numbers are essential for
communicating between man and the
computer. In applications requiring the
processing of a large volume of
inherently decimal input and output
data, the time for decimal-binary
conversion needed by a purely binary
computer may be significant. A slower
```

```
decimal adder may take less time than
a fast binary adder doing an addition
and two conversions. A careful review
of the significance of decimal and
binary addressing and both binary
and decimal data arithmetic,
supplemented by efficient
conversion instructions.
```

Here is the list of noun phrases extracted by *NPtool*. For the key, see Section 4.4.

```
ok: addition
ok: advantage
ok: application
ok: arithmetic speed
ok: binary address
ok: binary addressing
ok: binary and decimal data arithmetic
ok: binary computer
ok: binary number system
ok: careful review of the significance
    of decimal and binary addressing
ok: certain powerful operation
ok: computer
ok: decimal instruction format
ok: decimal number
ok: decimal representation
ok: decimal-binary conversion
ok: efficient conversion instruction
ok: fast binary adder
ok: high-performance, general-purpose
    computer
ok: greater compactness of binary number
ok: greater simplicity of a binary
    arithmetic unit
ok: instruction format
ok: man
ok: overall performance
ok: slower decimal adder
ok: time
ok: two conversion
ok: way
?:  communicating
?:  processing of a large volume of
    inherently decimal input
?:  processing of a large volume of
    inherently decimal input and
    output data
?:  output data
```